\documentclass[10pt,twocolumn,letterpaper]{article}

\usepackage{iccv}     %
\usepackage{times}
\usepackage{epsfig}
\usepackage{eso-pic}

\usepackage{graphicx}
\usepackage{amsmath}
\usepackage{amssymb}
\usepackage{booktabs}
\usepackage{capt-of}

\usepackage{multirow}

\usepackage{colortbl}  
\usepackage{xcolor}
\usepackage{array}
\definecolor{hollywoodcerise}{rgb}{0.96, 0.0, 0.63}
\definecolor{lasallegreen}{rgb}{0.03, 0.47, 0.19}
\definecolor{hanpurple}{rgb}{0.32, 0.09, 0.98}
\definecolor{green(pigment)}{rgb}{0.0, 0.65, 0.31}

\usepackage{hyperref}

\hypersetup{colorlinks,linkcolor={red},citecolor={hanpurple},urlcolor={red}}  

\usepackage{multirow}

\usepackage{colortbl}  
\usepackage{xcolor}
\usepackage{array}
\usepackage[hypcap = false]{caption}

\iccvfinalcopy 

\def\iccvPaperID{****} 

\ificcvfinal\fi

\begin{document}

\title{
Look at the Neighbor: Distortion-aware Unsupervised Domain Adaptation for Panoramic Semantic Segmentation
}

\author{Xu Zheng$^{1}$ \quad Tianbo Pan$^{1}$ \quad Yunhao Luo$^{3}$ \quad Lin Wang$^{1}$$^{,2}$\thanks{Corresponding author.}\\
$^{1}$AI Thrust, HKUST(GZ) \quad $^{2}$Dept. of CSE, HKUST \quad $^{3}$Brown University
\\
{\tt\small zhengxu128@gmail.com, tpan695@connect.hkust-gz.edu.cn, devinluo@gmail.com, linwang@ust.hk}
\\
\tt \small Project Page: \url{https://vlislab22.github.io/DATR/}
}

\twocolumn[{
\maketitle
\renewcommand\twocolumn[1][t!]{#1}%
\begin{center}
    \centering
    \vspace{-24pt}
    \includegraphics[width=\textwidth]{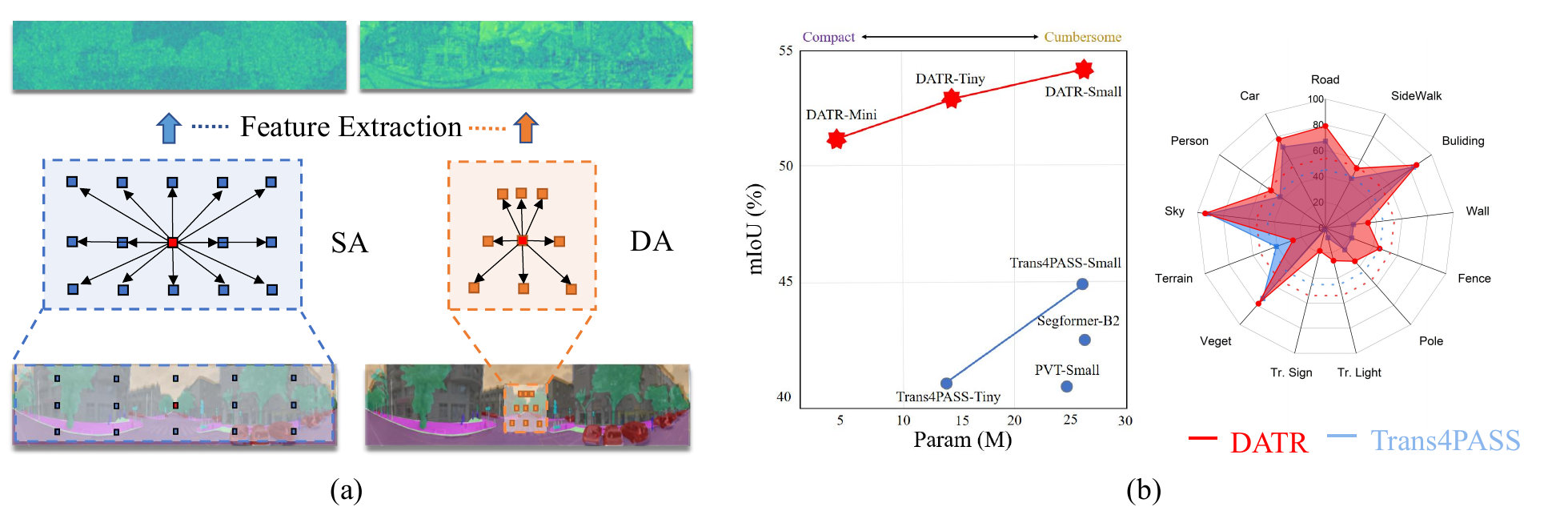}
    \end{center}
    \vspace{-8pt}
    \captionof{figure}{
    (a) We propose a novel UDA method that effectively addresses the distortion problems via the distortion-aware attention (DA) module that extracts more salient textural details than those of self-attention (SA) ~\cite{zhang2022bending}; 
    (b) Our method achieves more than \textbf{8$\%$} of mIoU than the SOTA method (with 24.98M parameters)~\cite{zhang2022bending} while taking only \textbf{4.64M} parameters on the Synthetic~\cite{zhang2022behind}-to-Real~\cite{densepass} scenarios. 
    }
    \label{teaser}
    \vspace{24pt}
}]
\renewcommand{\thefootnote}{} 
\footnotetext{$^*$Corresponding author.}
\begin{abstract}
Endeavors have been recently made to transfer knowledge from the labeled pinhole image domain to the unlabeled panoramic image domain via Unsupervised Domain Adaptation (UDA).
The aim is to tackle the domain gaps caused by the style disparities and distortion problem from the non-uniformly distributed pixels of equirectangular projection (ERP). Previous works typically focus on transferring knowledge based on geometric priors with specially designed multi-branch network architectures. As a result, considerable computational costs are induced, and meanwhile, their generalization abilities are profoundly hindered by the variation of distortion among pixels. In this paper, we find that the pixels' neighborhood regions of the ERP indeed introduce less distortion. Intuitively, we propose a novel UDA framework that can effectively address the distortion problems for panoramic semantic segmentation. In comparison, our method is simpler, easier to implement, and more computationally efficient.
Specifically, we propose distortion-aware attention (DA) capturing the neighboring pixel distribution without using any geometric constraints. 
Moreover, we propose a class-wise feature aggregation (CFA) module to iteratively update the feature representations with a memory bank. As such, the feature similarity between two domains can be consistently optimized. Extensive experiments show that our method achieves new state-of-the-art performance while remarkably reducing 80$\%$ parameters.

\end{abstract}

\section{Introduction}
\label{sec:intro}
The burgeoning demand for omnidirectional and dense scene understanding has stimulated the popularity of $360^\circ$ cameras, which pose much wider field-of-view (FoV) in the range of $360^\circ \times 180^\circ$ than the 2D images captured by pinhole cameras~\cite{aisurvey}. $360^\circ$ cameras deliver complete scene details either in the outdoor or indoor environment; therefore, research has been actively focused on panoramic semantic segmentation for the pixel-wise scene understanding of the intelligent systems, such as self-driving and augmented/virtual reality~\cite{sekkat2022comparative,xu2019semantic,yang2020omnisupervised}. 

Generally, $360^\circ$ images are projected into the 2D planar representations while preserving the omnidirectional information~\cite{yoon2022spheresr,disconmod, zheng2023both}, to be aligned with the existing pipelines~\cite{wang2023space, zheng2022transformer}. Equirectangular projection (ERP) is the most commonly used projection type. 
However, ERP images often suffer from the image distortion and object deformation~\cite{zhang2022behind}, caused by the non-uniformly distributed pixels. Also, the lack of precisely annotated datasets heavily impedes training effective panoramic semantic segmentation models.

For these reasons, research endeavors have been made to transfer knowledge from the labeled pinhole image domain to the unlabeled panoramic image domain via Unsupervised Domain Adaptation (UDA). It aims to tackle the domain gaps caused by intrinsic style disparities and inevitable distortion problems.
Typically, \cite{zhang2021transfer,p2pda,densepass,gu2021pit,zhang2022behind,zhang2022bending,tateno2018distortion} leverage the spatial geometric priors (\eg, convolution variants~\cite{tateno2018distortion} and attention-augmented components~\cite{zhang2022bending,zhang2022behind}) to address the distortion problems. 
However, these priors are essentially inadequate for the panoramic semantic segmentation; therefore, cumbersome, \ie, multi-branch network architectures~\cite{yang2020ds} are designed to reinforce the learning abilities.
Consequently, considerable computation costs are induced, and their generalization abilities are profoundly plagued by the variation of distortion among the pixels.

\begin{figure}[t!]
    \centering
    \includegraphics[width=0.8\columnwidth]{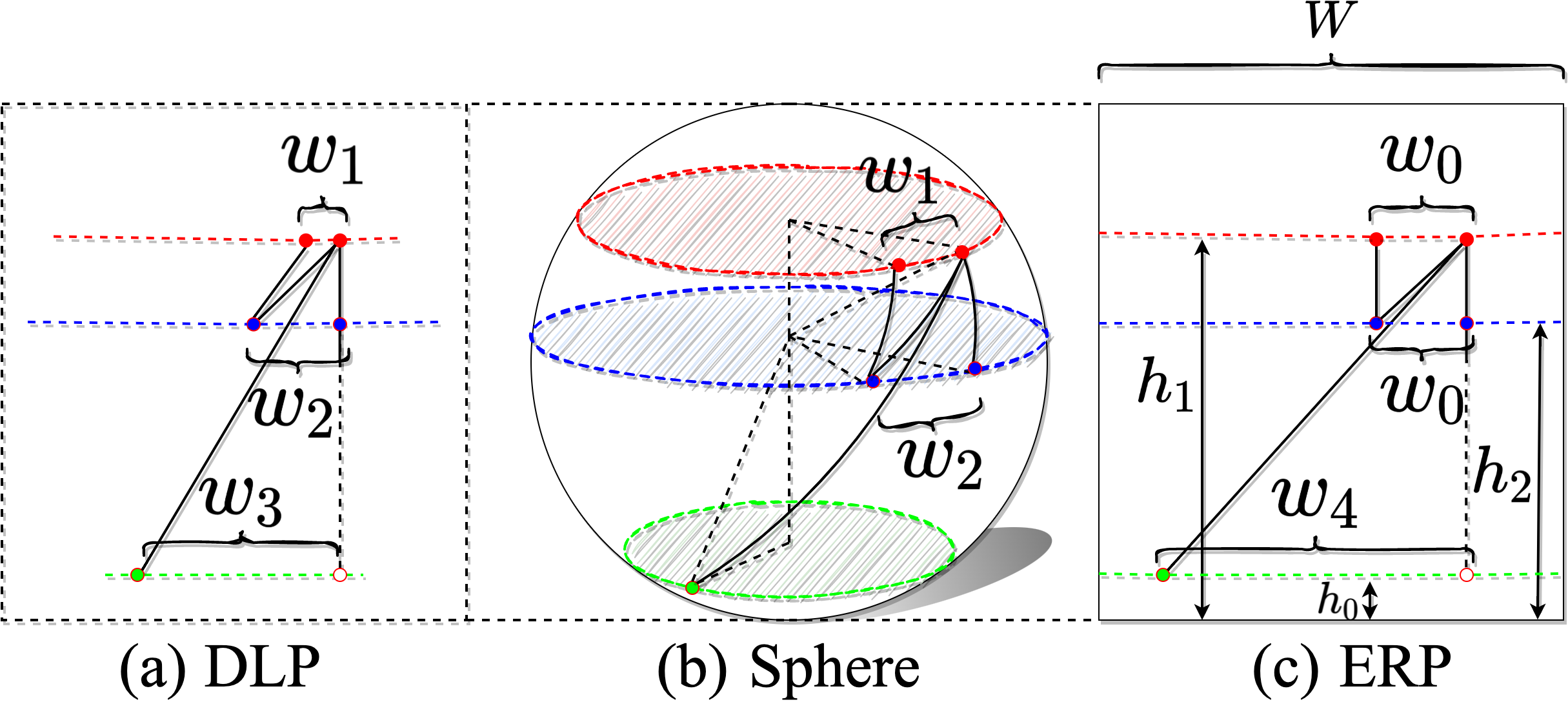}
    \caption{Direct Linear Projection (DLP) (a) and ERP (c) are two projection formats of the same spherical data (b).}
    \label{att_noa}
\end{figure}

In this paper, we find that the pixels' neighboring regions in the ERP indeed introduce less distortion.
As the ERP shuffles the equidistribution of spherical pixels, the distance (Fig.~\ref{att_noa} \textcolor{red}{(b)}) between any two pixels for a specific latitude of a $360^\circ$ image is different from that (Fig.~\ref{att_noa} \textcolor{red}{(c)}) of the ERP image (sphere-to-plane projection). 
As a result, it is easier to capture the positional distribution among the pixels by reducing the receptive field, which is more efficient in addressing distortion problems. Therefore, controlling the neighboring region size is crucial in balancing the trade-off between receptive field and distortion problems.

In light of this, we propose a novel UDA framework that can efficiently address the distortion problems for panoramic semantic segmentation. Compared with the state-of-the-art UDA methods~\cite{DAFormer,zhang2022bending,zhang2022behind,PCS,p2pda}, our method is simpler, easier to implement, and more computationally efficient.
Our method enjoys two key contributions. Firstly, we propose a novel distortion-aware attention (\textbf{DA}) module to capture the neighboring pixel distributions between domains (See Fig.~\ref{teaser} \textcolor{red}{(a)}).  
This is buttressed by a trainable relative positional encoding (RPE), which provides unique neighboring positional information.
We then build a hierarchically structured DA-based transformer (DATR) that aggregates the feature information from all layers.
In addition, we propose a class-wise feature aggregation (CFA) module that transfers knowledge of the extracted features between domains. It updates the class-wise feature centers with a memory bank and consistently optimizes the cross-domain feature similarity by iteratively updating class centers. 

We conduct extensive experiments for both the synthetic and real-world scenarios, including Cityscapes-to-DensePASS and Synthetic-to-DensePASS datasets. The results show that our framework surpasses the SOTA methods by +8.76$\%$ and +1.59$\%$ on the Synthetic-to-Real and Pinhole-to-Panoramic scenarios, respectively while taking only 20$\%$ parameters (See Fig.~\ref{teaser} \textcolor{red}{(b)}). 
In summary, our major contributions are three-fold: (\textbf{I})  Our work serves as the \textbf{first} attempt to address the distortion problems by capturing the neighboring pixel distributions. (\textbf{II}) We propose a DA module to capture the neighboring pixel distributions. (\textbf{III})  We propose a CFA module to iteratively transfer the cross-domain knowledge.
\section{Related Work}
\noindent\textbf{UDA for Panoramic Semantic Segmentation} 
can be divided into three major categories: adversarial learning, pseudo label generation, and feature prototype adaption. The first type of methods~\cite{Hoffman2018CyCADACA,Choi2019SelfEnsemblingWG,Sankaranarayanan2018LearningFS,Tsai2018LearningTA} tends to learn the domain invariance by conducting alignment from the image level~\cite{Hoffman2018CyCADACA,Li2019BidirectionalLF,Murez2018ImageTI}, feature level~\cite{Hoffman2018CyCADACA,Chen2019ProgressiveFA,Hoffman2016FCNsIT}, and output level~\cite{Luo2019TakingAC,MelasKyriazi2021PixMatchUD}.
The second type of methods generates pseudo labels for the target domain training and utilizes self-training to refine them. For example, ~\cite{Liu2021PanoSfMLearnerSM,Zhang2021DeepPanoContextP3,Wang2021DomainAS,Zhang2017CurriculumDA} conduct refinement by leveraging the guidance from an auxiliary task, \eg,~depth estimation~\cite{Liu2021PanoSfMLearnerSM}.
The third type of method, \eg, Mutual Prototype Adaption (MPA)~\cite{zhang2022bending}, aligns the feature embeddings with the prototypes obtained in the source and target domain individually. However, these approaches utilize the multi-stage training strategy and thus fail to correlate the features in each mini-batch.
Differently, we propose the CFA module to aggregate the class-wise prototypes and iteratively update them, promoting the prototypes to have a more holistic representation of the peculiarity of domains.\\

\noindent\textbf{Distortion Problems of ERP.}
Previous works on alleviating the distortion problems are with the manner of deformable kernel~\cite{Su2017LearningSC,Coors_2018_ECCV} and designing adaptable CNN~\cite{tateno2018distortion,zhao2018distortion} according to the geometrical priors of the sphere.
Particularly, ~\cite{zhang2022behind} adaptively adjusts the receptive field during patch embedding to better retain the semantic consistency and considers the distortion problems during the feature parsing process. However, this method is inevitably inefficient due to the large receptive field (See Fig.~\ref{teaser} \textcolor{red}{(b)}).
Another way is designing distortion-aware neural networks~\cite{zhang2022bending,zhang2022behind}. Deformable components, \eg, Deformable Patch Embedding (DPE) and Deformable MLP (DMLP)~\cite{zhang2022bending}, are widely explored for panoramic semantic segmentation as they can help to learn the prior knowledge of panorama characteristics when patchifying the input data. Though the performance is significantly improved, the abilities of data generalization are limited, and mostly rely on prior geometry knowledge.
Differently, we find that the neighboring region of ERP indeed introduces less distortion, benefiting the generalization to the variance of pixel distribution. Therefore, we propose the DA module to address the distortion problem with much fewer parameters.\\
 

\noindent \textbf{Self-attention (SA)} is defined as a dot product operation on query, key, and value sequence ~\cite{attentionisallyouneed}. Dosovitskiy \etal ~\cite{dosovitskiy2020image} first proposed to utilize the SA on image patches in the vision field. More recently, abundant variants of attention paradigms~\cite{liu2021swin,xie2021segformer,qin2021fcanet,chu2021twins} are proposed to tackle vision problems. For panoramic semantic segmentation, Multi-Head Self-Attention (MHSA)~\cite{attentionisallyouneed} and Efficient Self-Attention (ESA)~\cite{xie2021segformer} are extensively applied to capture the long-range dependencies of $360^\circ$ images. However, MHSA and ESA are inadequate in alleviating the distortion problem between pixels caused by the global feature extraction strategy. By contrast, our work shares a different spirit by focusing on the neighboring pixels, and accordingly the DA module is proposed to reduce the distortion problems by capturing the distinct pixels' distribution between domains.
\section{Method}
\subsection{Theoretical Analysis of ERP Distortion }
\label{fewerdis}
The most notable advantage of ERP is its ease of operation due to its projection strategy, which is a completely linear transformation~\cite{aisurvey}. However, ERP shuffles the equidistribution of sphere pixels, resulting in varying distances between pixels at different latitudes before and after ERP, as shown in Fig.~\ref{att_noa} \textcolor{red}{(a)}, \textcolor{red}{(b)}, and \textcolor{red}{(c)}, where the distances between pixels $w_1$, $w_2$, and $w_0$ are different.
\begin{equation}
    \begin{aligned}
    & w_0 = W/n, 
    w_1 = \frac{2 \pi}{n}\sqrt{h_1 (\frac{W}{\pi}-h_1)}, \\
    & w_2 = \frac{2 \pi}{n}\sqrt{h_2 (\frac{W}{\pi}-h_2)}, 
    w_0 > w_2 > w_1,
\end{aligned}
\end{equation}
where $n$ is the number of sampling pixels at each latitude, $W$ is the width of ERP and $h_1$, $h_2$ is the height of red and blue pixels.
The non-uniform sampling density of spherical data caused by the ERP projection can lead to distortion, as the sampling density is different at the poles and equator.

\noindent \textbf{Our idea:}  
We quantify the lateral distortion as the difference in distance between the pixels on a distinct projection type. By contrast, the vertical distortion is evenly distributed. Intuitively, we formulate the distortion coefficient $Dis$ between the red and green points in Fig.~\ref{att_noa} as:
\begin{equation}
    Dis = w_4 - w_3 = \frac{n'}{n}(W - 2\pi \sqrt{h_0(\frac{W}{\pi}-h_o)} ), 
\label{long}
\end{equation}
where $n'$ is the number of pixels between the red and green points in the latitude coordinate. 
\begin{figure}[]
    \centering
    \includegraphics[width=0.8\columnwidth]{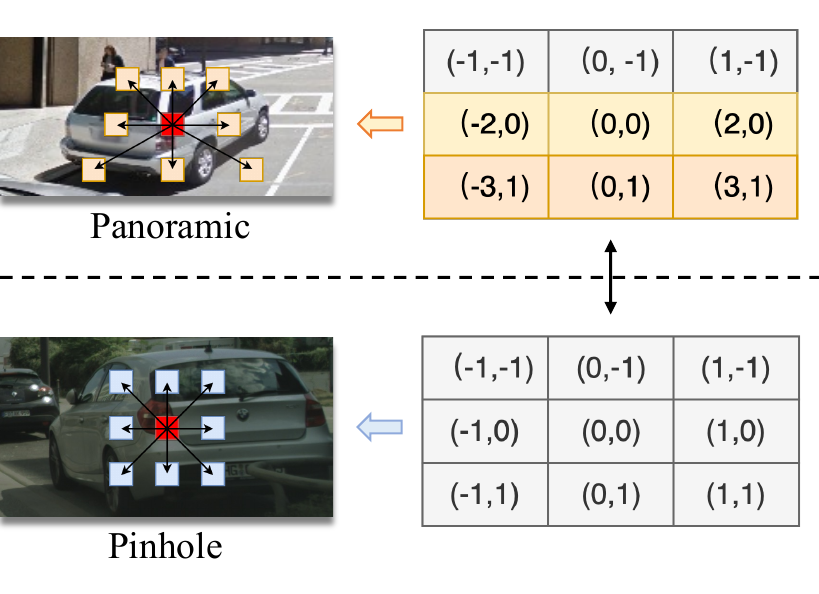}
    \caption{Illustration of the proposed RPE.}
    \label{RPE}
\end{figure}
Eq.\ref{long} demonstrates that $Dis$ increases with $n'$, indicating that a smaller value of $n'$ leads to less distortion. This observation leads us to propose addressing the distortion problem in UDA for panoramic semantic segmentation from a neighboring perspective. By minimizing the receptive field of the UDA network model, we can more effectively capture the pixel's distribution and address the distortion problem. Based on this theoretical analysis, we propose a distortion-aware attention (DA) module in Sec.\ref{da_module_sec} to mitigate this issue. \textit{More detailed analysis see supplmat.}
\begin{figure*}[t!]
    \centering
    \includegraphics[width=\textwidth]{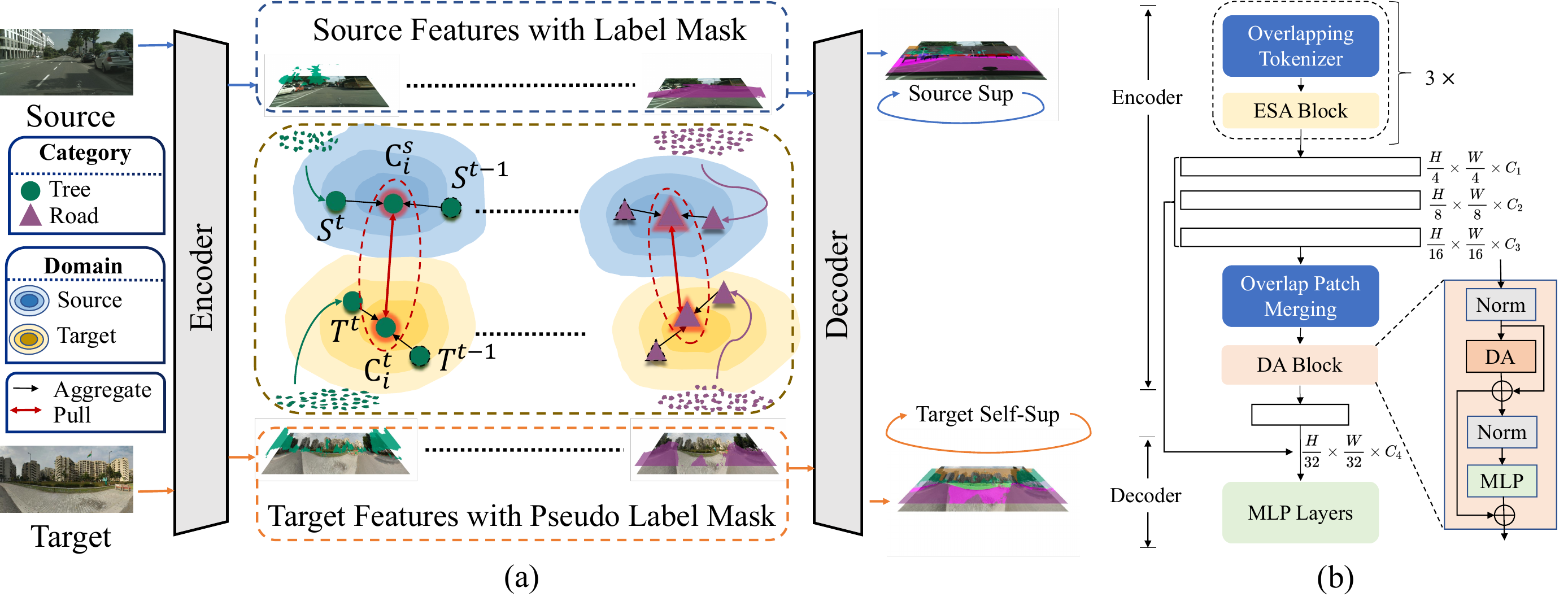}
    \caption{(a) Illustration of the proposed CFA module; (b) The architecture of our proposed DA-based transformer.}
    \label{Framework}
\end{figure*}
\subsection{Distortion-aware UDA Framework}
\subsubsection{Distortion-aware Attention (DA)}
\label{da_module_sec}
We now describe how to tackle the distortion problem with the proposed DA module. 
It captures the pixel distribution in the neighborhood region. 
As shown in Fig.~\ref{teaser} \textcolor{red}{(a)}, the pixels in the DA only look at the neighboring pixels. To get the intuition, let's first review the MHSA~\cite{attentionisallyouneed}, which can be formulated as:
\begin{equation}
\label{mhsa}
    Attention(Q,K,V) = Softmax(\frac{QK^{\top}}{\sqrt{d_{head}}})V,
\end{equation}
where $Q$, $K$, and $V$ are the linear projections of the whole input sequence.
Based on our observation, we propose to learn UDA networks efficiently without strong geometric constraints on the pixel's neighboring regions. To achieve this, we narrow the receptive field and focus more on the local distribution of neighboring pixels to minimize the distortion coefficient, as defined in Eq.~\ref{long}. To be specific, we denote the nearest pixels of $p_{i,j}$ in the $H \times W$ neighborhood region as $P_{i,j}$, where $H$ and $W$ represent the height and width of the region, respectively. Thus, the DA can be defined as:
\begin{equation}
\label{NOA}
    DA(p_{i,j}) = Softmax(\frac{Q_{p_{i,j}}K^\top_{P_{i,j}}}{scale} )V_{P_{i,j}},
\end{equation}
where $Q$, $K$, and $V$ are the linear projections of the corresponding input, the neighboring pixels, and have the same dimensions $N \times C$, and  $N = H \times W$. \\

\noindent \textbf{Relative Positional Encoding (RPE).} Self-attention-based models suffer from the inability to capture the order of input tokens or patches~\cite{rpe}. To address this limitation for panoramic semantic segmentation, it is crucial to incorporate explicit representations of positional information. Previous MHSA-based methods~\cite{zhang2022bending,zhang2022behind} leverage the geometric properties of the ERP to design specific fixed positional encoding. However, these methods rely heavily on given priors and lack adaptive distortion-aware abilities. In contrast, our RPE provides local neighboring positional information to alleviate this problem. As a result, our DA module can ignore the resolution difference between training and inference.
As depicted in Fig.\ref{RPE}, it is evident that the distribution of neighboring pixels in the pinhole image (in blue) differs from that in the panoramic image (in yellow). To address this issue, we propose a trainable Relative Positional Encoding (RPE) for DA. Specifically, RPE consists of a set of trainable embedding vectors, initialized with a uniform distribution, that represents the position encoding of each pixel. The RPE captures the distorted pixel distribution of ERP after adaptation and remains fixed during inference. This encoding method captures the distribution of different neighboring pixels and enables the UDA model to address the domain gap resulting from the inevitable distortion. The encoding vectors are embedded into the DA module, allowing us to reformulate Eq.\ref{NOA} as follows:
\begin{equation}
    DA(p_{i,j}) = Softmax(\frac{Q_{p_{i,j}}K^\top_{P_{i,j}}}{scale} )(V_{P_{i,j}} + RPE),
\end{equation}
where the RPE is our proposed positional encoding.
\subsubsection{DA Block Design}
We now present the design of the DA block for distortion-aware feature extraction. In contrast to prior methods that use a transformer backbone~\cite{zhang2022bending,zhang2022behind}, our DA block incorporates local position information through RPE, rather than a feed-forward network. As depicted in Figure~\ref{Framework} \textcolor{red}{(b)}, our DA block consists of normalization and MLP layers, making the feature extraction process completely convolution-free. \textit{Additional details of the DA block are in the supplmat.}
\subsubsection{DA-based Transformer}
With the proposed DA block, we introduce a DA-based transformer model that is partially inspired by ViT~\cite{dosovitskiy2020image}. Differently, our DA-based transformer model is tailored to address distortion problems and serves as a backbone for UDA in panoramic semantic segmentation. Our model adopts an encoder-decoder structure without extra components, resulting in fewer parameters than multi-branch models such as PASTS ~\cite{PASTS} (\textbf{4.64M vs. 614M}). In experiments, our DATR outperforms existing complex models~\cite{xie2021segformer,zhang2022bending, DAFormer,zhang2022behind,wang2021pyramid} (see Tab.\ref{csdp_perclass}) and is more robust to adaptation stability and efficiency (see Tab.\ref{abDATR}). More specifically, our model shares a different spirit from prior methods as it is tailored to specifically address the distortion problem in panoramic segmentation. Further details of our DA-based transformer model can be found in the supplmat.\\

\noindent \textbf{Encoder.}
As shown in Fig.~\ref{Framework} \textcolor{red}{(b)}, our DA-based transformer backbone can generate multi-scale features, including coarse shallow layers and fine-grained deep layers.
To preserve the local continuity around the neighboring pixels, we use the overlapping patch merging of~\cite{xie2021segformer}. The patch size $K$, stride $S$, and padding size $P$ are set to 7/4/3 and 3/2/1 for the model with different scales.
Concretely, given an input image with a resolution of $H \times W$, the hierarchical feature maps $F_i$ have the resolutions of 
$\frac{H}{2^{i+1}} \times \frac{W}{2^{i+1}} \times C_i$,
where $i$ $\in$ ${1,2,3,4}$.
For a better trade-off between receptive fields and distortion-aware abilities (See Tab.~\ref{locations}), the first three layers of our model are based on efficient self-attention, and the deepest layer is based on our DA block.
To reduce the computational cost of ESA ($O(N^2)$), we adopt the sequence reduction process utilized in \cite{wang2021pyramid,xie2021segformer}. \\

\noindent \textbf{Decoder.}
As depicted in Fig.~\ref{Framework} \textcolor{red}{(b)}, the MLP decoder takes multi-scale features $F_i$ from the encoder as inputs, and the channel dimensions are aligned. Then $F_i$ are up-sampled to $F_4$'s size and concatenated and fused together. Finally, the last MLP layer takes the fused features to predict the segmentation confidence maps. 

\subsection{Class-wise Feature Aggregation (CFA)}
Given the source (\ie, synthetic or pinhole images) domain dataset with a set of annotated images $X_s$ = {($x^s$, $y^s$), $x^s$ $\in$ $R^{H \times W \times 3}$, $y^s$ $\in$ ${0,1}^{H \times W \times K}$} and the target (\ie, panoramic images) dataset $D_t$ = {($x^t$), $x^t$ $\in$ $R^{H \times W \times 3}$} without corresponding labels, the objective of UDA is to transfer knowledge from the source domain to the target domain with $K$ shared classes. Our model is first trained with the source domain data $D_s$ using the segmentation loss:
\begin{equation}
    \mathcal{L}_{SEG} = -\sum_{h,w,k=1}^{H,W,K}y^s_{(h,w,k)log(p^s_{h,w,k})},
\end{equation}
where $p^s_{(i,j,k)}$ is the prediction of the source image pixel $x^s_{(h,w)}$ as the $k$-th class. To adopt our model to the target domain data, we utilize the predictions of target images as the pseudo labels and perform self-supervised (SS) training. The calculation and generation metric is:
\begin{equation}
    \hat{y}^t_{(h,w,k)} = 1_{k \doteq argmax(p^t_{h,w,:}) }.
\end{equation}
With the pseudo labels, our model is optimized by:
\begin{equation}
    \mathcal{L}^t_{SS} = -\sum_{h,w,k=1}^{H,W,K}\hat{y} ^t_{(h,w,k)}log(p^t_{h,w,k})  
\end{equation}

We propose a novel class-wise feature aggregation (CFA) module to transfer knowledge from extracted features between domains, as illustrated in Fig.\ref{Framework} \textit{\textcolor{red}{(a)}. Our CFA module differs from prior UDA methods\cite{zhang2022bending,zhang2022behind} in two distinct ways: (a) it iteratively aggregates class-wise features and updates feature centers, thereby aligning them directly between the two domains; (b) hard pseudo-labels $\hat{y} ^t_{(h,w,k)}$ are softened in the feature space and used for prediction, enabling full leveraging of knowledge from the source domain.}
Specifically, we utilize the pseudo labels $\hat{y} ^t_{(h,w,k)}$ as the class-wise masks in the high-level feature space. 
Given the target domain feature's $i$th class center $T_i^t$ in $t$th iteration and the class center from the source domain feature is $S_i^t$, the extracted features are first masked by the pseudo labels and then are projected to class-wise feature centers $S^t$ $\backslash$ $T^t$. Note that $S^t$ $\backslash$ $T^t$ is mixed with the corresponding feature center $S^{t-1}$ $\backslash$ $T^{t-1}$ from the last iteration (mini-batch). We leverage this iterative mixing strategy for the feature center to make the centers (a.k.a, prototypes) more robust: 
\begin{equation}
\scalebox{0.9}{$\displaystyle
    C^s_i = (1 - \frac{1}{e})S_i^{(t-1)} + \frac{1}{e}S_i^t,\\
    C^t_i = (1 - \frac{1}{e})T_i^{(t-1)} + \frac{1}{e}T_i^t,$}
\end{equation}
where the $C^s_i$ and $C^t_i$ are the class centers of the source and target images, and $e$ is the current epoch number. 
The same class-wise feature centers are pushed together by the Mean Squared Error (MSE) to mimic the class-wise knowledge between domains:
\begin{equation}
    \mathcal{L}_f = \frac{1}{num}\sum_{i \in C}(C^s_i - C^t_i)^2,
\end{equation}
where $num$ is the total category number.  
\begin{figure*}[t!]
    \centering
    \includegraphics[width=\textwidth]{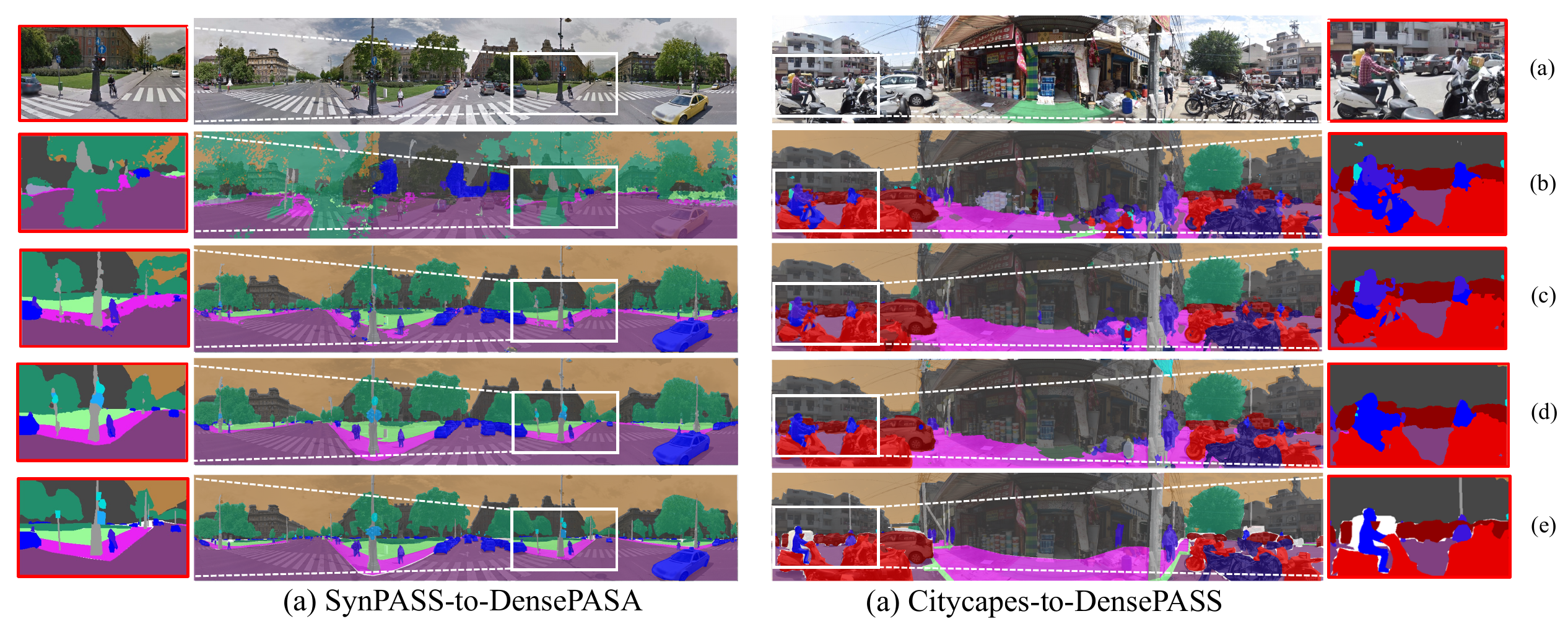}
    \caption{Example visualization results from DensePASS test set. (a) Input, (b) Fully supervised Segformer-B1 without domain adaptation~\cite{xie2021segformer}, (c) Trans4PASS-S~\cite{zhang2022bending}, (d) DATR-S, and (e) Ground truth.}
    \label{vis_results}
\end{figure*}
\section{Experiments}
To evaluate the effectiveness of our proposed UDA framework, we conducted extensive experiments on both real-world and synthetic datasets, which include Cityscapes~\cite{Cityscapes}, DensePASS~\cite{densepass}, and SynPASS~\cite{zhang2022behind}. We designed three instance models of the DATR, namely DATR-Mini, DATR-Tiny, and DATR-Small, with the same architecture but different sizes. DATR-Mini is the most efficient model, with the least computation cost during training and testing (4.64M). On the other hand, DATR-Small is the most powerful model with the best performance.
\subsection{Datasets and Implementation Details}
\noindent{\textbf{Cityscapes~\cite{Cityscapes}}} is a widely used dataset of urban street images captured from 50 different cities, with precise annotations of 19 categories. The official split includes 2975 images for training and 500 images for validation. In this paper, we use the official training set as the source data.\\

\noindent{\textbf{SynPASS~\cite{zhang2022behind}}} is a dataset of 9080 synthetic panoramic images annotated with 22 categories. The official training, validation, and test sets contain 5700, 1690, and 1690 images, respectively. We use the overlapping 13 classes between SynPASS and DensePASS datasets for training and testing.\\

\noindent{\textbf{DensePASS~\cite{densepass}}} is a 360-degree image dataset collected from 40 cities. The official training and test sets contain 2000 and 100 images, respectively, and are annotated with the same 19 classes as Cityscapes.\\

\noindent{\textbf{Implementation Details.}}
We train all the models using 4 NVIDIA GPUs, with an initial learning rate of $5\times10^{-5}$. The learning rate is scheduled using a polynomial strategy with a power of 0.9. We use the AdamW optimizer with an epsilon of $1\times10^{-8}$ and weight decay of $1\times10^{-4}$. The resolutions and data augmentations for the training and test images are kept the same as those used in ~\cite{zhang2022bending}.
\subsection{Experimental Results}
\begin{table}[t!]
\centering
\setlength{\tabcolsep}{1mm}
\resizebox{0.49\textwidth}{!}{
\begin{tabular}{cccccc}
\toprule
Network                      & Backbone       & CS    & DP    & Gap & Param   \\ \midrule
\multirow{2}{*}{DeepLabv3+~\cite{deeplabv3+}}  & ResNet-50~\cite{he2016deep}    & 79.81 & 29.79 & 50.02 & 39.76\\ 
                             & ResNet-101~\cite{he2016deep}   & 80.89 & 32.69 & 48.20 & 58.75 \\ \midrule
\multirow{2}{*}{Segformer~\cite{xie2021segformer}}   & MiT-B1  & 78.85 & 36.98 & 41.87 & 14.72 \\
                             & MiT-B2     & 81.32 & 41.23 & 40.09 & 25.76 \\ \midrule
\multirow{2}{*}{Trans4PASS~\cite{zhang2022bending}}  & Trans4PASS-T  & 80.54 & 41.56 & 38.98 & 13.95 \\ 
                             & Trans4PASS-S  & 81.17 & 42.47 & 38.70 & 24.98 \\ \midrule
\multirow{2}{*}{Trans4PASS+~\cite{zhang2022behind}} & Trans4PASS-T+ & 79.92 & 41.33 & 38.59 &13.95 \\ 
                             & Trans4PASS-S+  & 81.76 & 42.35 & 39.41 &24.98\\ \midrule
\rowcolor{gray!10}        & DATR-M          & 75.23      & 38.48     & \textbf{36.75}   & 4.64  \\ 
                       \rowcolor{gray!20}  \multirow{-1}{*}{DATR}    & DATR-T         &  79.01     & 42.22     & \textbf{36.79}  & 14.72   \\ 
                       \rowcolor{gray!30}      & DATR-S         &  79.98      & 47.55      & \textbf{32.43}   & 25.76   \\ \bottomrule
\end{tabular}}
\caption{Performance gaps of some SOTA CNN-based and transformer-based panoramic semantic segmentation models. The test size on DensePASS (DP) is 400 $\times$ 2048. All models are only trained with the Cityscapes dataset without the adaption module.}
\label{gap}
\end{table}

\begin{table*}[t!]
\centering
\setlength{\tabcolsep}{0.5mm}
\resizebox{\textwidth}{!}{
\begin{tabular}{ccccccccccccccccccccc}
\toprule
Method             & mIoU  &Road  & S.W. & Build. & Wall  & Fence & Pole  & Tr.L. & Tr.S. & Veget. & Terr. & Sky   & Persin & Rider & Car   & Truck & Bus   & Train & M.C. & B.C. \\ \midrule
ERFNet ~\cite{romera2017erfnet}& 16.65 & 63.59 & 18.22    & 47.01   & 9.45 & 12.79 & 17.00  & 8.12  & 6.41 & 34.24 & 10.15 & 18.43 & 4.96  & 2.31  & 46.03 & 3.19  & 0.59  & 0.00  &8.30  & 5.55   \\
PASS(ERFNet) ~\cite{yang2019pass}     & 23.66 & 67.84 & 28.75    & 59.69    & 19.96 & 29.41 & 8.26  & 4.54          & 8.07         & 64.96      & 13.75   & 33.50 & 12.87  & 3.17  & 48.26 & 2.17  & 0.82  & 0.29  & 23.76      & 19.46   \\ 
Omni-sup(ECANet) ~\cite{yang2020omnisupervised}   & 43.02 & 81.60 & 19.46    & 81.00    & 32.02 & 39.47 & 25.54 & 3.85          & 17.38        & 79.01      & 39.75   & 94.60 & 46.39  & 12.98 & 81.96 & 49.25 & 28.29 & 0.00  & 55.36      & 29.47   \\ 
P2PDA(Adversarial) ~\cite{p2pda} & 41.99 & 70.21 & 30.24    & 78.44    & 26.72 & 28.44 & 14.02 & 11.67         & 5.79         & 68.54      & 38.20   & 85.97 & 28.14  & 0.00  & 70.36 & 60.49 & 38.90 & 77.80 & 39.85      & 24.02   \\ 
PCS ~\cite{PCS}& 53.83 & 78.10 & 46.24 & 86.24  & 30.33 &45.78 & 34.04  & 22.74  & 13.00 & 79.98 & 33.07 & 93.44 & 47.69  & 22.53  & 79.20 & 61.59  & 67.09  & 83.26  & 58.68  & 39.80   \\
DAFormer~\cite{DAFormer} & 54.67 & 73.75 & 27.34 & 86.35 & 35.88 & 45.56 & 36.28 & 25.53 &10.65 &79.87 & 41.64 &94.74 & 49.69 & 25.15 & 77.70 &63.06 & 65.61 & 86.68 & 65.12 & 48.13
\\
Trans4PASS-T ~\cite{zhang2022bending} & 53.18 & 78.13 & 41.19    & 85.93    & 29.88 & 37.02  & 32.54 & 21.59         & 18.94        & 78.67      &45.20   &93.88 & 48.54  & 16.91 & 79.58 & 65.33 & 55.76 & 84.63 & 59.05      & 37.61   \\ 
Trans4PASS-S ~\cite{zhang2022bending}  & 55.22 & 78.38 & 41.58    & 86.48 & 31.54 & 45.54  & 33.92 & 22.96 & 18.27 & 79.40 & 41.07 & 93.82 & 48.85  & 23.36 &81.02 &67.31 &69.53 & 86.13 & 60.85 & 39.09   \\ 
\midrule
\rowcolor{gray!10} Ours w/ DATR-M  &52.90 &78.71 &48.43 &86.92 &34.92 &43.90 &33.43 &22.39 &17.15 &78.55 &28.38 &93.72 &52.08 &13.24 &77.92 &56.73 &59.53 &\textbf{93.98} &51.12 &34.06 \\ 
\rowcolor{gray!20} Ours w/ DATR-T  &54.60 &79.43 &49.70 &87.39 &37.91 &44.85 &35.06 &25.16 &19.33 &78.73 &25.75 &93.60 &53.52 &20.20 &78.07 &60.43 &55.82 &91.11 &67.03 &34.32 \\ 
\rowcolor{gray!30} Ours w/ DATR-S &\textbf{56.81} &\textbf{80.63} &\textbf{51.77} &\textbf{87.80} &\textbf{44.94} &43.73 &\textbf{37.23} &\textbf{25.66} &\textbf{21.00}& 78.61 &26.68 &93.77 &\textbf{54.62} &\textbf{29.50} &80.03 &\textbf{67.35} &63.75 &87.67 &\textbf{67.57} &37.10  \\
\bottomrule
\end{tabular}}
\caption{Per-class results of the SOTA panoramic image semantic segmentation methods on DensePASS test set.}
\label{csdp_perclass}
\end{table*}

\begin{table*}[]
\centering
\small
\resizebox{\textwidth}{!}{
\begin{tabular}{cccccccccccccccc}
\toprule
Method                       & Backbone      & mIoU & Road & S.Walk & Build. & Wall & Fence & Pole & Tr.L. & Tr.S. & Veget. & Terrain & Sky & Person & Car \\ \midrule
\multirow{4}{*}{Source-only} & PVT-S &38.74 &55.39 &36.87 &80.84 &19.72 &15.18 & 8.04 &5.39 &2.17 &72.91 &32.01 &90.81 &26.76 &57.40\\  
& Trans4PASS+-S &43.17 &73.72 &43.31 &79.88 &19.29 &16.07 &20.02 &8.83 &1.72 &67.84 &31.06 &86.05 &44.77 &68.58\\
& DATR-M &32.37 &62.48 &17.00 &74.55 &10.66 &7.17 &11.48 &3.37 &0.38 &60.21 &17.56 &81.99 &26.38 &47.55\\ 
& DATR-T &33.82 &61.33 &16.77 &75.44 &14.64 &7.65 &15.11 &4.10 &1.30 &64.06 &15.25 &82.72 &26.13 &55.21 \\ 
& DATR-S &35.15 &60.43 &13.57 &76.69 &18.35 &5.88 &17.33 &3.44 &2.62 &62.68 &19.54 &83.58 &34.30 &58.56 \\ \midrule
\multirow{5}{*}{MPA~\cite{zhang2022behind}}         & PVT-S  &40.90 &70.78 &42.47 &82.13 &22.79 &10.74 &13.54 &1.27 &0.30 &71.15 &33.03 &89.69 &29.07 &64.73\\ 
& Trans4PASS+-S &45.29 &67.28 &43.48 &83.18 &22.02 &21.98 &22.72 &7.86 &1.52 &73.12 &40.65 &91.36 &42.69 &70.87\\
& DATR-M & 48.24 & 77.05 & 46.43 & 83.80 & 25.16 & 35.25 &26.20 &19.12 & 12.54 & 77.93 & 23.79 & 94.23 & 38.04 & 67.59\\
& DATR-T & 52.11 & 78.40 & 52.10 & 85.04 & 31.52 & 42.44 & 30.11 & 22.78 & 15.01 & 77.50 & 27.96 & 93.76 & 47.08 & 73.80\\ 
& DATR-S & 52.76 & 78.33 & 52.70 & 85.15 & 30.69 & 42.59 & 32.19 & 24.20 & 17.90 & 77.72 & 27.24 & 93.86 & 47.98 & 75.34\\ 
\midrule
\rowcolor{gray!10}        
& DATR-M &51.04 &77.62 &49.03 &84.58 &28.15 &39.70 &30.34 &23.83 & 15.95 &78.23 &24.74 &93.73 & 44.14  & 74.45\\ 
\rowcolor{gray!20} \multirow{-1}{*}{CFA (ours)} & DATR-T & 53.23 & 79.09 & 52.92 & 85.51 & 32.02 & 42.90 & 31.56 & 27.17 & 17.14 & 77.87 & 28.71 & 93.72 & 48.16 & 75.26\\ 
\rowcolor{gray!30}& DATR-S &\textbf{54.05} & \textbf{79.07} &\textbf{52.28} &\textbf{85.98} &\textbf{33.38} &\textbf{45.02} &\textbf{34.47} &\textbf{26.15} &\textbf{18.27} &\textbf{78.21} &26.99 &\textbf{94.02} &\textbf{51.21} &\textbf{77.62} \\
\bottomrule
\end{tabular}}
\caption{We evaluate all the UDA methods with various backbones on the SynPASS-to-DensePASS scenario, the overlapped 13 classes (DensePASS13) of two datasets are used to test the UDA performance. }
\label{abDATR}
\end{table*}
We first evaluate our proposed framework on Cityscapes (pinhole)-to-DensePASS (panorama) datasets. Tab.~\ref{gap} presents the mIoU results evaluated on the Cityscapes and DensePASS test set. Our DATR without any adaption module has the least performance drop (36.75$\%$) compared with the SOTA segmentation methods~\cite{deeplabv3+, xie2021segformer, zhang2022bending, zhang2022behind}. Importantly, our DATR-M is much more compact than the prior models, with only 4.64M parameters, and has a competitive and even superior performance.
In Tab.~\ref{csdp_perclass}, we compare DATR against previous SOTA approaches, including PASS~\cite{yang2019pass}, Omni-sup~\cite{yang2020omnisupervised}, P2PDA~\cite{p2pda}, PCS~\cite{PCS} and Trans4PASS~\cite{zhang2022bending}. Among these methods, though DAFormer~\cite{DAFormer} serves as the SOTA transformer-based DA method, Trans4PASS achieves better results. Yet our DATR-S reaches the mIoU of 56.81$\%$ on the DensePASS test set while outperforming the DAFomer and Trans4PASS by mIoU increment of +2.14$\%$ and +1.59$\%$, respectively. Evidently, large improvements (over 3$\%$) have been obtained on sidewalk, wall, pole, \etc, which are challenging yet pivotal categories in practical applications. 

We further evaluate our method on SynPASS (synthetic)-to-DensePASS (real-world) datasets. Though synthetic panoramas' spatial distributions and content are closer to the real panoramic data, the previous methods struggle on several categories, \eg., traffic light and traffic sign, which strongly rely on texture cues. The main reasons are two folds: 1) the simulated traffic elements in the synthetic data do not present diverse textures and details, which aggravate the domain gap; 2) prior UDA methods fail to capture the regional pixel-wise correspondence. As shown in 
Tab.~\ref{abDATR}, all of the variants of our DATR consistently outperform previous SOTA methods. Especially, our DATR-S achieves dramatically mIoU increment on the most challenging categories, including \textit{Fence} (+23.04$\uparrow$), \textit{Pole} (+11.75$\uparrow$), \textit{Tr.Light} (+18.29$\uparrow$), \textit{Tr.Sign} (+16.75$\uparrow$), \textit{Person} (+8.52$\uparrow$), and \textit{Car} (+6.75$\uparrow$). This is also demonstrated in Fig.~\ref{vis_results}. Overall, even the most compact model, DATR-M, achieves +5.75$\%$ mIoU increment than the prior SOTA method Trans4PASS+-S~\cite{zhang2022behind}(24.98M) with only 4.64M parameters. This indicates that our DATR model brings largely enhanced domain adaptation performances in synthetic-to-real scenarios, buttressed by the proposed DA and CFA modules.
\section{Ablation Study and Analysis}
\noindent{\textbf{Design Choices for DA and ESA.}}
We first demonstrate the superiority and necessity for the simultaneous usage of ESA and DA by conducting ablation experiments. We set different experiment settings, including DA at different structure locations. As we can see in Tab.~\ref{locations}, only using ESA suffers unsatisfactory performance. Meanwhile, using DA in the shallow layers also leads to poor performance due to the limited local receptive fields of the DA module. By contrast, utilizing DA at the deep layer brings the largest performance gain (+2.03$\%$ $\uparrow$). It validates that using ESA and DA together in the proper sequence makes DATR achieves better distortion-aware abilities.\\

\noindent{\textbf{Effectiveness of Relative Positional Encoding.}}
We now study the effectiveness of the proposed RPE in DA. We test three positional encoding strategies including absolute positional encoding (APE), relative positional encoding (RPE), and without any positional encoding (without PE). Our DATR-M achieves 41.56$\%$, 51.04$\%$, and 36.98$\%$ mIoU with APE, RPE and without PE, respectively. This indicates the significant effectiveness of our proposed RPE in conferring distortion-aware abilities to models.\\

\begin{figure*}[t]
    \centering
    \includegraphics[width=\textwidth]{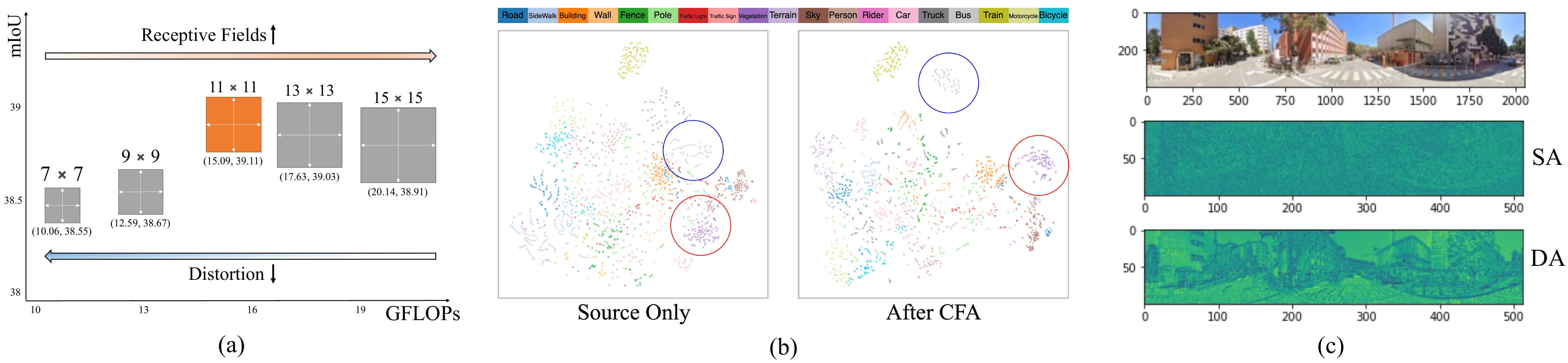}
    \caption{(a) Ablation study of different Neighborhood region sizes with DATR-M framework on the DensePASS test set. (b) t-SNE visualizations with different modules on the DensePASS test set with DATR-T. (c) Visualization of the extracted features from panoramic images by Self-Attention (SA) and Distortion-aware Attention (DA).}
    \label{ablation}
\end{figure*}

\noindent{\textbf{Ablation of UDA Module.}}
To verify the effectiveness of our proposed UDA module, we conduct experiments on two benchmarks with different modules. As shown in Tab.~\ref{udaab}, evidently, models trained with CFA achieve over 6$\%$ mIoU increment on Cityscapes-to-DensePASS than the models only trained with the source (Cityscapes) data. Moreover, our CFA gives more than 9$\%$ mIoU improvement to all variants of DATR, which indicates the effectiveness of our CFA and the structure superiority of DATR. As for \textit{SynPASS to DensePASS}, our Self-Sup strategy and CFA achieve more than 10$\%$ mIoU increment. \\

\noindent \textbf{Generality of CFA.} In SynPASS-to-DensePASS scenario, our CFA achieves 48.33$\%$ and 44.97$\%$ and 48.96$\%$ mIOU with Trans4PASS-S, PVT-S and DAFormer. Compared with the MPA Tab.~\ref{abDATR}, our CFA achieves higher UDA performance by 3.04$\%$, 4.07$\%$, and 3.71$\%$ than MPA, respectively. \textit{The results confirm that CFA provides a better learning signal to diverse models}. \\

\noindent{\textbf{Effectiveness of DATR.}}
We evaluate the generalization capacity of our introduced DATR by applying it in other domain adaptation methods, \eg, MPA~\cite{zhang2022bending}. As shown in Tab.~\ref{abDATR}, our DATR consistently achieves superior performance than PVT~\cite{wang2021pyramid} and Trans4PASS+~\cite{zhang2022behind}. Specifically, our most efficient instance model DATR-M (4.64M) outperforms PVT-S (24.5M) and Trans4PASS+-S (24.98M) by +7.34$\%$ and 2.95$\%$ mIoU increments, respectively. Evidently, large improvements have been made by our DATR-S model, which achieves +11.86$\%$ and +7.47$\%$ compared 
with PVT-S and Trans4PASS+-S. This indicates the superior generalization capacity of our proposed DATR, facing domain shift behind 360$^\circ$ and pinhole imagery.
\section{Discussion}
\noindent{\textbf{Significance of Neighboring Region Size.}}
The DA's cover size has a significant impact on the performance of our proposed DATR. When the neighborhood region covers the entire input, DA behaves like self-attention. However, using a smaller neighborhood size results in simpler distortion but smaller receptive fields, which can be fatal for semantic segmentation tasks. We conducted experiments with different neighborhood region sizes ranging from $7 \times 7$ to $15 \times 15$ and found that much smaller or larger region sizes did not bring satisfactory results. To balance GFLOPs and performance, we choose the region size as $11 \times 11$.\\

\noindent{\textbf{Performance of small objects.}}
Our proposed DATR demonstrates its capability of performing well on relatively small objects, such as traffic lights, through the use of DA blocks with narrowed receptive fields of attention. This is evidenced by the performance gains on such objects, with an increase of 18.29$\%$ as presented in Tab.~\ref{abDATR}. Moreover, DATR also shows reasonable performance gains on larger objects, such as the sky (+2.66$\%$) and buildings (+2.80$\%$). \\

\noindent{\textbf{Rationality of DA and ESA.}}
To demonstrate the segmentation performance on panoramic images, we conduct ablation experiments of training a single layer of DA and ESA. As shown in Fig.~\ref{ablation} \textcolor{red}{(c)}, our proposed DA shows more salient textural details than ESA \wrt the visualization of the extracted features, showing the superiority of our DA.\\
\begin{table}[t!]
\centering
\setlength{\tabcolsep}{1mm}
\resizebox{0.49\textwidth}{!}{
\begin{tabular}{c|c|c|c|c|c|c|c}
\toprule
Structure & $\circ$ $\circ$ $\circ$ $\circ$    & $\star$ $\circ$ $\circ$ $\circ$     & $\circ$ $\star$ $\circ$ $\circ$     & $\circ$ $\circ$ $\star$ $\circ$     & $\circ$ $\circ$ $\circ$ $\star$     & $\star$ $\circ$ $\star$ $\circ$     & $\circ$ $\star$ $\circ$ $\star$     \\ \midrule
mIoU     & 36.45 & 37.56 & 38.07 & 38.29 & \textbf{38.48} & 37.15 & 38.17 \\ \midrule
$\Delta$     & - & +1.11$\uparrow$ & +1.62$\uparrow$ & +1.84$\uparrow$ & \
\textbf{+2.03$\uparrow$} & +0.70$\uparrow$ & +1.72$\uparrow$ \\ \bottomrule
\end{tabular}}
\caption{Ablation study of the proper location of our proposed DA block. $\star$: DA block, $\circ$: ESA block.}
\label{locations}
\end{table}

\begin{table}[t!]
\setlength{\tabcolsep}{0.5mm}
\resizebox{0.49\textwidth}{!}{
\begin{tabular}{cccc|cccc}
\toprule
\multicolumn{4}{c}{Cityscapes -\textgreater DensePASS}     & \multicolumn{4}{c}{SynPASS -\textgreater DensePASS}      \\ \midrule
Backbone                & Method & mIoU  & $\delta$  & Backbone                & Method & mIoU & $\delta$ \\ \midrule
\multirow{3}{*}{DATR-M} & Source     & 38.48 & -      & \multirow{3}{*}{DATR-M} & Source     & 33.01     & -     \\ 
                        & SS        & 45.71 & +7.23$\uparrow$  &                         & SS        & 45.52     & +12.51$\uparrow$      \\  
                        & CFA             & 52.90 & +14.42$\uparrow$ &                         & CFA             & 51.04     & +18.03$\uparrow$      \\ \midrule
\multirow{3}{*}{DATR-T} & Source     & 42.22 & -      & \multirow{3}{*}{DATR-T} & Source     & 33.98     & -     \\ 
                        & SS        & 48.27 & +6.05$\uparrow$  &                         & SS       & 47.72     & +13.74$\uparrow$      \\  
                        & CFA             & 54.60 & +12.40$\uparrow$ &                         & CFA             & 53.23     & +19.25$\uparrow$      \\ \midrule
\multirow{3}{*}{DATR-S} & Source     & 47.55 & -      & \multirow{3}{*}{DATR-S} & Source     & 36.74     & -     \\  
                        & SS        & 54.96 & +7.41$\uparrow$   &                         & SS        & 50.08     & +13.86$\uparrow$      \\ 
                        & CFA             & 56.81 & +9.26$\uparrow$   &                         & CFA             & 54.05     & +17.31$\uparrow$      \\ \bottomrule
\end{tabular}}
\caption{Ablation study of different module combinations on two public benchmarks.}
\label{udaab}
\end{table}

\noindent{\textbf{Effectiveness and rationality of CFA.}}
To demonstrate the effectiveness and rationality of our proposed CFA, we conducted ablation experiments using our proposed DATR model. The results in Tab.\ref{abDATR} show that our CFA consistently outperforms the previous state-of-the-art method MPA\cite{zhang2022bending}. With all variants of our proposed DATR model, our CFA achieved the highest mIoU, surpassing MPA by 1.29$\%$, 1.12$\%$, and 2.80$\%$ mIoU, respectively. Additionally, we used t-SNE visualization to compare the extracted features before and after incorporating our CFA on the DensePASS test set. As shown in Fig.~\ref{ablation} \textcolor{red}{(b)}, CFA provides holistic class-wise feature aggregation, resulting in more closely tied features within the same class, such as the red and blue circles. \\

\noindent{\textbf{Iterative Strategy in CFA.}}
To ablate the impact of our proposed iterative class-wise feature aggregation module, we substitute the iterative strategy with non-iterative and linear iterative methods to demonstrate the effectiveness of our iterative strategy. We conduct experiments with DATR-M by non-iterative and linear iterative strategy, the mIoU results are 50.43$\%$ and 50.53$\%$, respectively. Our iterative update makes the class center more robust and holistic and achieves 51.04$\%$ mIoU with the same backbone.\\

\noindent \textbf{Analysis of failure classes.}
Our DATR, which uses DA blocks in the deepest layer, performs well on relatively small objects, such as traffic lights (+18.29\%) shown in Tab.~\ref{abDATR}. It also achieves reasonable performance gains on large objects, such as sky (+2.66\%) and building (+2.80\%). Consequently, the performance of each class is related to the neighboring size of DA, as shown in Fig.~\ref{ablation}\textcolor{red}{(a)}.\\

\noindent \textbf{Unrestricted resolutions.} Our DATR computes attention in a fixed neighborhood region, which is not largely affected by input sizes. We test different resolutions during inference, and the performance variation among different input sizes did not exceed 10\%, while compared models~\cite{zhang2022bending, zhang2022behind}, fluctuate over 20\% (See Tab.4 in the suppl.).\\
\section{Conclusion and Future Work}
In this paper, we found that the pixels' neighborhood regions of ERP indeed introduce less distortion. Based on our observation, we proposed a novel distortion-aware attention (DA) module that focuses on capturing the neighboring pixel distribution, buttressed by a trainable relative positional encoding (RPE). We further built a unified backbone model for panoramic semantic segmentation. Moreover, we proposed a class-wise feature aggregation (CFA) module to iteratively update the features with a memory bank. As such, we consistently optimized the feature similarity between domains. Our proposed framework significantly outperformed the SOTA UDA methods with an order of magnitude fewer parameters.\\

\noindent \textbf{Limitation and future work:} In this paper, the UDA is performed on a single source dataset to a single target dataset. 
It would be worth exploring the multi-source domain adaptation for panoramic semantic segmentation. Moreover, we will explore how to utilize the other projections of the sphere data to facilitate knowledge transfer.\\ 

\noindent \textbf{Acknowledgement:} This work was supported by the National Natural Science Foundation of China (NSF) under Grant No.NSFC22FYT45.
\clearpage
{\small
\bibliographystyle{ieee_fullname}
\bibliography{egbib}

\begin{thebibliography}{10}\itemsep=-1pt

\bibitem{aisurvey}
Hao Ai, Zidong Cao, Jinjing Zhu, Haotian Bai, Yucheng Chen, and Ling Wang.
\newblock Deep learning for omnidirectional vision: A survey and new
  perspectives.
\newblock {\em arXiv preprint arXiv:2205.10468}, 2022.

\bibitem{Chen2019ProgressiveFA}
Chaoqi Chen, Weiping Xie, Tingyang Xu, Wenbing Huang, Yu Rong, Xinghao Ding,
  Yue Huang, and Junzhou Huang.
\newblock Progressive feature alignment for unsupervised domain adaptation.
\newblock {\em 2019 IEEE/CVF Conference on Computer Vision and Pattern
  Recognition (CVPR)}, pages 627--636, 2019.

\bibitem{deeplabv3+}
Liang-Chieh Chen, Yukun Zhu, George Papandreou, Florian Schroff, and Hartwig
  Adam.
\newblock Encoder-decoder with atrous separable convolution for semantic image
  segmentation.
\newblock In {\em Proceedings of the European conference on computer vision
  (ECCV)}, pages 801--818, 2018.

\bibitem{Choi2019SelfEnsemblingWG}
Jaehoon Choi, Taekyung Kim, and Changick Kim.
\newblock Self-ensembling with gan-based data augmentation for domain
  adaptation in semantic segmentation.
\newblock {\em 2019 IEEE/CVF International Conference on Computer Vision
  (ICCV)}, pages 6829--6839, 2019.

\bibitem{chu2021twins}
Xiangxiang Chu, Zhi Tian, Yuqing Wang, Bo Zhang, Haibing Ren, Xiaolin Wei,
  Huaxia Xia, and Chunhua Shen.
\newblock Twins: Revisiting the design of spatial attention in vision
  transformers.
\newblock {\em Advances in Neural Information Processing Systems},
  34:9355--9366, 2021.

\bibitem{Coors_2018_ECCV}
Benjamin Coors, Alexandru~Paul Condurache, and Andreas Geiger.
\newblock Spherenet: Learning spherical representations for detection and
  classification in omnidirectional images.
\newblock In {\em Proceedings of the European Conference on Computer Vision
  (ECCV)}, September 2018.

\bibitem{Cityscapes}
Marius Cordts, Mohamed Omran, Sebastian Ramos, Timo Rehfeld, Markus Enzweiler,
  Rodrigo Benenson, Uwe Franke, Stefan Roth, and Bernt Schiele.
\newblock The cityscapes dataset for semantic urban scene understanding.
\newblock In {\em Proc. of the IEEE Conference on Computer Vision and Pattern
  Recognition (CVPR)}, 2016.

\bibitem{dosovitskiy2020image}
Alexey Dosovitskiy, Lucas Beyer, Alexander Kolesnikov, Dirk Weissenborn,
  Xiaohua Zhai, Thomas Unterthiner, Mostafa Dehghani, Matthias Minderer, Georg
  Heigold, Sylvain Gelly, et~al.
\newblock An image is worth 16x16 words: Transformers for image recognition at
  scale.
\newblock In {\em International Conference on Learning Representations}, 2020.

\bibitem{gu2021pit}
Qiqi Gu, Qianyu Zhou, Minghao Xu, Zhengyang Feng, Guangliang Cheng, Xuequan Lu,
  Jianping Shi, and Lizhuang Ma.
\newblock Pit: Position-invariant transform for cross-fov domain adaptation.
\newblock In {\em Proceedings of the IEEE/CVF International Conference on
  Computer Vision}, pages 8761--8770, 2021.

\bibitem{he2016deep}
Kaiming He, Xiangyu Zhang, Shaoqing Ren, and Jian Sun.
\newblock Deep residual learning for image recognition.
\newblock In {\em Proceedings of the IEEE conference on computer vision and
  pattern recognition}, pages 770--778, 2016.

\bibitem{Hoffman2018CyCADACA}
Judy Hoffman, Eric Tzeng, Taesung Park, Jun-Yan Zhu, Phillip Isola, Kate
  Saenko, Alexei~A. Efros, and Trevor Darrell.
\newblock Cycada: Cycle-consistent adversarial domain adaptation.
\newblock In {\em ICML}, 2018.

\bibitem{Hoffman2016FCNsIT}
Judy Hoffman, Dequan Wang, Fisher Yu, and Trevor Darrell.
\newblock Fcns in the wild: Pixel-level adversarial and constraint-based
  adaptation.
\newblock {\em ArXiv}, abs/1612.02649, 2016.

\bibitem{DAFormer}
Lukas Hoyer, Dengxin Dai, and Luc~Van Gool.
\newblock Daformer: Improving network architectures and training strategies for
  domain-adaptive semantic segmentation.
\newblock In {\em {IEEE/CVF} Conference on Computer Vision and Pattern
  Recognition, {CVPR} 2022, New Orleans, LA, USA, June 18-24, 2022}, pages
  9914--9925. {IEEE}, 2022.

\bibitem{disconmod}
Xing Hu, Yi An, Cheng Shao, and Huosheng Hu.
\newblock Distortion convolution module for semantic segmentation of panoramic
  images based on the image-forming principle.
\newblock {\em IEEE Transactions on Instrumentation and Measurement}, 71:1--12,
  2022.

\bibitem{PASTS}
Jihyun Kim, Somi Jeong, and Kwanghoon Sohn.
\newblock Pasts: Toward effective distilling transformer for panoramic semantic
  segmentation.
\newblock In {\em 2022 IEEE International Conference on Image Processing
  (ICIP)}, pages 2881--2885, 2022.

\bibitem{Li2019BidirectionalLF}
Yunsheng Li, Lu Yuan, and Nuno Vasconcelos.
\newblock Bidirectional learning for domain adaptation of semantic
  segmentation.
\newblock {\em 2019 IEEE/CVF Conference on Computer Vision and Pattern
  Recognition (CVPR)}, pages 6929--6938, 2019.

\bibitem{Liu2021PanoSfMLearnerSM}
Mengyi Liu, Shuhui Wang, Yulan Guo, Yuan He, and Hui Xue.
\newblock Pano-sfmlearner: Self-supervised multi-task learning of depth and
  semantics in panoramic videos.
\newblock {\em IEEE Signal Processing Letters}, 28:832--836, 2021.

\bibitem{liu2021swin}
Ze Liu, Yutong Lin, Yue Cao, Han Hu, Yixuan Wei, Zheng Zhang, Stephen Lin, and
  Baining Guo.
\newblock Swin transformer: Hierarchical vision transformer using shifted
  windows.
\newblock In {\em Proceedings of the IEEE/CVF International Conference on
  Computer Vision}, pages 10012--10022, 2021.

\bibitem{Luo2019TakingAC}
Yawei Luo, Liang Zheng, Tao Guan, Junqing Yu, and Yi Yang.
\newblock Taking a closer look at domain shift: Category-level adversaries for
  semantics consistent domain adaptation.
\newblock {\em 2019 IEEE/CVF Conference on Computer Vision and Pattern
  Recognition (CVPR)}, pages 2502--2511, 2019.

\bibitem{densepass}
Chaoxiang Ma, Jiaming Zhang, Kailun Yang, Alina Roitberg, and Rainer
  Stiefelhagen.
\newblock Densepass: Dense panoramic semantic segmentation via unsupervised
  domain adaptation with attention-augmented context exchange.
\newblock In {\em 2021 IEEE International Intelligent Transportation Systems
  Conference (ITSC)}, pages 2766--2772. IEEE, 2021.

\bibitem{MelasKyriazi2021PixMatchUD}
Luke Melas-Kyriazi and Arjun~K. Manrai.
\newblock Pixmatch: Unsupervised domain adaptation via pixelwise consistency
  training.
\newblock {\em 2021 IEEE/CVF Conference on Computer Vision and Pattern
  Recognition (CVPR)}, pages 12430--12440, 2021.

\bibitem{Murez2018ImageTI}
Zak Murez, Soheil Kolouri, David~J. Kriegman, Ravi Ramamoorthi, and Kyungnam
  Kim.
\newblock Image to image translation for domain adaptation.
\newblock {\em 2018 IEEE/CVF Conference on Computer Vision and Pattern
  Recognition}, pages 4500--4509, 2018.

\bibitem{qin2021fcanet}
Zequn Qin, Pengyi Zhang, Fei Wu, and Xi Li.
\newblock Fcanet: Frequency channel attention networks.
\newblock In {\em Proceedings of the IEEE/CVF international conference on
  computer vision}, pages 783--792, 2021.

\bibitem{romera2017erfnet}
Eduardo Romera, Jos{\'e}~M Alvarez, Luis~M Bergasa, and Roberto Arroyo.
\newblock Erfnet: Efficient residual factorized convnet for real-time semantic
  segmentation.
\newblock {\em IEEE Transactions on Intelligent Transportation Systems},
  19(1):263--272, 2017.

\bibitem{Sankaranarayanan2018LearningFS}
Swami Sankaranarayanan, Yogesh Balaji, Arpit Jain, Ser-Nam Lim, and Rama
  Chellappa.
\newblock Learning from synthetic data: Addressing domain shift for semantic
  segmentation.
\newblock {\em 2018 IEEE/CVF Conference on Computer Vision and Pattern
  Recognition}, pages 3752--3761, 2018.

\bibitem{sekkat2022comparative}
Ahmed~Rida Sekkat, Yohan Dupuis, Paul Honeine, and Pascal Vasseur.
\newblock A comparative study of semantic segmentation of omnidirectional
  images from a motorcycle perspective.
\newblock {\em Scientific Reports}, 12(1):1--14, 2022.

\bibitem{Su2017LearningSC}
Yu-Chuan Su and Kristen Grauman.
\newblock Learning spherical convolution for fast features from 360° imagery.
\newblock In {\em NIPS}, 2017.

\bibitem{tateno2018distortion}
Keisuke Tateno, Nassir Navab, and Federico Tombari.
\newblock Distortion-aware convolutional filters for dense prediction in
  panoramic images.
\newblock In {\em Proceedings of the European Conference on Computer Vision
  (ECCV)}, pages 707--722, 2018.

\bibitem{Tsai2018LearningTA}
Yi-Hsuan Tsai, Wei-Chih Hung, Samuel Schulter, Kihyuk Sohn, Ming-Hsuan Yang,
  and Manmohan Chandraker.
\newblock Learning to adapt structured output space for semantic segmentation.
\newblock {\em 2018 IEEE/CVF Conference on Computer Vision and Pattern
  Recognition}, pages 7472--7481, 2018.

\bibitem{attentionisallyouneed}
Ashish Vaswani, Noam Shazeer, Niki Parmar, Jakob Uszkoreit, Llion Jones,
  Aidan~N. Gomez, Lukasz Kaiser, and Illia Polosukhin.
\newblock Attention is all you need.
\newblock In Isabelle Guyon, Ulrike von Luxburg, Samy Bengio, Hanna~M. Wallach,
  Rob Fergus, S.~V.~N. Vishwanathan, and Roman Garnett, editors, {\em Advances
  in Neural Information Processing Systems 30: Annual Conference on Neural
  Information Processing Systems 2017, December 4-9, 2017, Long Beach, CA,
  {USA}}, pages 5998--6008, 2017.

\bibitem{wang2023space}
Changqi Wang, Haoyu Xie, Yuhui Yuan, Chong Fu, and Xiangyu Yue.
\newblock Space engage: Collaborative space supervision for contrastive-based
  semi-supervised semantic segmentation, 2023.

\bibitem{Wang2021DomainAS}
Qin Wang, Dengxin Dai, Lukas Hoyer, Olga Fink, and Luc~Van Gool.
\newblock Domain adaptive semantic segmentation with self-supervised depth
  estimation.
\newblock {\em 2021 IEEE/CVF International Conference on Computer Vision
  (ICCV)}, pages 8495--8505, 2021.

\bibitem{wang2021pyramid}
Wenhai Wang, Enze Xie, Xiang Li, Deng-Ping Fan, Kaitao Song, Ding Liang, Tong
  Lu, Ping Luo, and Ling Shao.
\newblock Pyramid vision transformer: A versatile backbone for dense prediction
  without convolutions.
\newblock In {\em Proceedings of the IEEE/CVF International Conference on
  Computer Vision}, pages 568--578, 2021.

\bibitem{rpe}
Kan Wu, Houwen Peng, Minghao Chen, Jianlong Fu, and Hongyang Chao.
\newblock Rethinking and improving relative position encoding for vision
  transformer.
\newblock In {\em Proceedings of the IEEE/CVF International Conference on
  Computer Vision}, pages 10033--10041, 2021.

\bibitem{xie2021segformer}
Enze Xie, Wenhai Wang, Zhiding Yu, Anima Anandkumar, Jose~M Alvarez, and Ping
  Luo.
\newblock Segformer: Simple and efficient design for semantic segmentation with
  transformers.
\newblock {\em Advances in Neural Information Processing Systems},
  34:12077--12090, 2021.

\bibitem{xu2019semantic}
Yuanyou Xu, Kaiwei Wang, Kailun Yang, Dongming Sun, and Jia Fu.
\newblock Semantic segmentation of panoramic images using a synthetic dataset.
\newblock In {\em Artificial Intelligence and Machine Learning in Defense
  Applications}, volume 11169, pages 90--104. SPIE, 2019.

\bibitem{yang2019pass}
Kailun Yang, Xinxin Hu, Luis~M Bergasa, Eduardo Romera, and Kaiwei Wang.
\newblock Pass: Panoramic annular semantic segmentation.
\newblock {\em IEEE Transactions on Intelligent Transportation Systems},
  21(10):4171--4185, 2019.

\bibitem{yang2020ds}
Kailun Yang, Xinxin Hu, Hao Chen, Kaite Xiang, Kaiwei Wang, and Rainer
  Stiefelhagen.
\newblock Ds-pass: Detail-sensitive panoramic annular semantic segmentation
  through swaftnet for surrounding sensing.
\newblock In {\em 2020 IEEE Intelligent Vehicles Symposium (IV)}, pages
  457--464. IEEE, 2020.

\bibitem{yang2020omnisupervised}
Kailun Yang, Xinxin Hu, Yicheng Fang, Kaiwei Wang, and Rainer Stiefelhagen.
\newblock Omnisupervised omnidirectional semantic segmentation.
\newblock {\em IEEE Transactions on Intelligent Transportation Systems}, 2020.

\bibitem{yoon2022spheresr}
Youngho Yoon, Inchul Chung, Lin Wang, and Kuk-Jin Yoon.
\newblock Spheresr: 360deg image super-resolution with arbitrary projection via
  continuous spherical image representation.
\newblock In {\em Proceedings of the IEEE/CVF Conference on Computer Vision and
  Pattern Recognition}, pages 5677--5686, 2022.

\bibitem{PCS}
Xiangyu Yue, Zangwei Zheng, Shanghang Zhang, Yang Gao, Trevor Darrell, Kurt
  Keutzer, and Alberto~Sangiovanni Vincentelli.
\newblock Prototypical cross-domain self-supervised learning for few-shot
  unsupervised domain adaptation.
\newblock In {\em Proceedings of the IEEE/CVF Conference on Computer Vision and
  Pattern Recognition}, pages 13834--13844, 2021.

\bibitem{Zhang2021DeepPanoContextP3}
Cheng Zhang, Zhaopeng Cui, Cai Chen, Shuaicheng Liu, Bing Zeng, Hujun Bao, and
  Yinda Zhang.
\newblock Deeppanocontext: Panoramic 3d scene understanding with holistic scene
  context graph and relation-based optimization.
\newblock {\em 2021 IEEE/CVF International Conference on Computer Vision
  (ICCV)}, pages 12612--12621, 2021.

\bibitem{zhang2021transfer}
Jiaming Zhang, Chaoxiang Ma, Kailun Yang, Alina Roitberg, Kunyu Peng, and
  Rainer Stiefelhagen.
\newblock Transfer beyond the field of view: Dense panoramic semantic
  segmentation via unsupervised domain adaptation.
\newblock {\em IEEE Transactions on Intelligent Transportation Systems}, 2021.

\bibitem{p2pda}
Jiaming Zhang, Chaoxiang Ma, Kailun Yang, Alina Roitberg, Kunyu Peng, and
  Rainer Stiefelhagen.
\newblock Transfer beyond the field of view: Dense panoramic semantic
  segmentation via unsupervised domain adaptation.
\newblock {\em IEEE Transactions on Intelligent Transportation Systems}, 2021.

\bibitem{zhang2022bending}
Jiaming Zhang, Kailun Yang, Chaoxiang Ma, Simon Rei{\ss}, Kunyu Peng, and
  Rainer Stiefelhagen.
\newblock Bending reality: Distortion-aware transformers for adapting to
  panoramic semantic segmentation.
\newblock In {\em Proceedings of the IEEE/CVF Conference on Computer Vision and
  Pattern Recognition}, pages 16917--16927, 2022.

\bibitem{zhang2022behind}
Jiaming Zhang, Kailun Yang, Hao Shi, Simon Rei{\ss}, Kunyu Peng, Chaoxiang Ma,
  Haodong Fu, Kaiwei Wang, and Rainer Stiefelhagen.
\newblock Behind every domain there is a shift: Adapting distortion-aware
  vision transformers for panoramic semantic segmentation.
\newblock {\em arXiv preprint arXiv:2207.11860}, 2022.

\bibitem{Zhang2017CurriculumDA}
Yang Zhang, Philip David, and Boqing Gong.
\newblock Curriculum domain adaptation for semantic segmentation of urban
  scenes.
\newblock {\em 2017 IEEE International Conference on Computer Vision (ICCV)},
  pages 2039--2049, 2017.

\bibitem{zhao2018distortion}
Qiang Zhao, Chen Zhu, Feng Dai, Yike Ma, Guoqing Jin, and Yongdong Zhang.
\newblock Distortion-aware cnns for spherical images.
\newblock In {\em IJCAI}, pages 1198--1204, 2018.

\bibitem{zheng2022transformer}
Xu Zheng, Yunhao Luo, Hao Wang, Chong Fu, and Lin Wang.
\newblock Transformer-cnn cohort: Semi-supervised semantic segmentation by the
  best of both students.
\newblock {\em arXiv preprint arXiv:2209.02178}, 2022.

\bibitem{zheng2023both}
Xu Zheng, Jinjing Zhu, Yexin Liu, Zidong Cao, Chong Fu, and Lin Wang.
\newblock Both style and distortion matter: Dual-path unsupervised domain
  adaptation for panoramic semantic segmentation.
\newblock In {\em Proceedings of the IEEE/CVF Conference on Computer Vision and
  Pattern Recognition}, pages 1285--1295, 2023.

\end{thebibliography}
}

\end{document}